\documentclass[10pt,twocolumn,letterpaper]{article}

\usepackage{cvpr}
\usepackage{times}
\usepackage{epsfig}
\usepackage{graphicx}
\usepackage{amsmath}
\usepackage{amssymb}
\usepackage{tabulary}
\usepackage{multirow}

\usepackage{soul}

\usepackage[bookmarks=false,colorlinks=true,linkcolor=black,citecolor=black,filecolor=black,urlcolor=black]{hyperref}
\usepackage[british,UKenglish,USenglish,english,american]{babel}

\cvprfinalcopy 

\newcommand{\ve}[1]{\mathbf{#1}} 

\newcommand{\tabincell}[2]{\begin{tabular}{@{}#1@{}}#2\end{tabular}}

\renewcommand\arraystretch{1.2}

\begin{document}

\title{Deep Residual Learning for Image Recognition}

\author{Kaiming He \qquad Xiangyu Zhang \qquad Shaoqing Ren \qquad Jian Sun \\
\large Microsoft Research \vspace{-.2em}\\
\normalsize
\{kahe,~v-xiangz,~v-shren,~jiansun\}@microsoft.com
}

\maketitle

\begin{abstract}
\vspace{-.5em}
Deeper neural networks are more difficult to train. We present a residual learning framework to ease the training of networks that are substantially deeper than those used previously. We explicitly reformulate the layers as learning residual functions with reference to the layer inputs, instead of learning unreferenced functions. We provide comprehensive empirical evidence showing that these residual networks are easier to optimize, and can gain accuracy from considerably increased depth.
On the ImageNet dataset we evaluate residual nets with a depth of up to 152 layers---8$\times$ deeper than VGG nets \cite{Simonyan2015} but still having lower complexity.
An ensemble of these residual nets achieves 3.57\% error on the ImageNet \emph{test} set. This result won the 1st place on the ILSVRC 2015 classification task.
We also present analysis on CIFAR-10 with 100 and 1000 layers.

The depth of representations is of central importance for many visual recognition tasks. Solely due to our extremely deep representations, we obtain a 28\% relative improvement on the COCO object detection dataset. Deep residual nets are foundations of our submissions to ILSVRC \& COCO 2015 competitions\footnote{\fontsize{7.6pt}{1em}\selectfont \url{http://image-net.org/challenges/LSVRC/2015/} and \url{http://mscoco.org/dataset/\#detections-challenge2015}.}, where we also won the 1st places on the tasks of ImageNet detection, ImageNet localization, COCO detection, and COCO segmentation.
\end{abstract}




\vspace{-1em}
\section{Introduction}
\label{sec:intro}

Deep convolutional neural networks \cite{LeCun1989,Krizhevsky2012} have led to a series of breakthroughs for image classification \cite{Krizhevsky2012,Zeiler2014,Sermanet2014}. Deep networks naturally integrate low/mid/high-level features \cite{Zeiler2014} and classifiers in an end-to-end multi-layer fashion, and the ``levels'' of features can be enriched by the number of stacked layers (depth).
Recent evidence \cite{Simonyan2015,Szegedy2015} reveals that network depth is of crucial importance, and the leading results \cite{Simonyan2015,Szegedy2015,He2015,Ioffe2015} on the challenging ImageNet dataset \cite{Russakovsky2014} all exploit ``very deep'' \cite{Simonyan2015} models, with a depth of sixteen \cite{Simonyan2015} to thirty \cite{Ioffe2015}. Many other nontrivial visual recognition tasks \cite{Girshick2014,He2014,Girshick2015,Ren2015,Long2015} have also greatly benefited from very deep models.

Driven by the significance of depth, a question arises: \emph{Is learning better networks as easy
as stacking more layers?}
An obstacle to answering this question was the notorious problem of vanishing/exploding gradients \cite{Bengio1994,Glorot2010}, which hamper convergence from the beginning. This problem, however, has been largely addressed by normalized initialization \cite{LeCun1998,Glorot2010,Saxe2013,He2015} and intermediate normalization layers \cite{Ioffe2015}, which enable networks with tens of layers to start converging for stochastic gradient descent (SGD) with backpropagation \cite{LeCun1989}.

\begin{figure}[t]
\begin{center}
\includegraphics[width=1.0\linewidth]{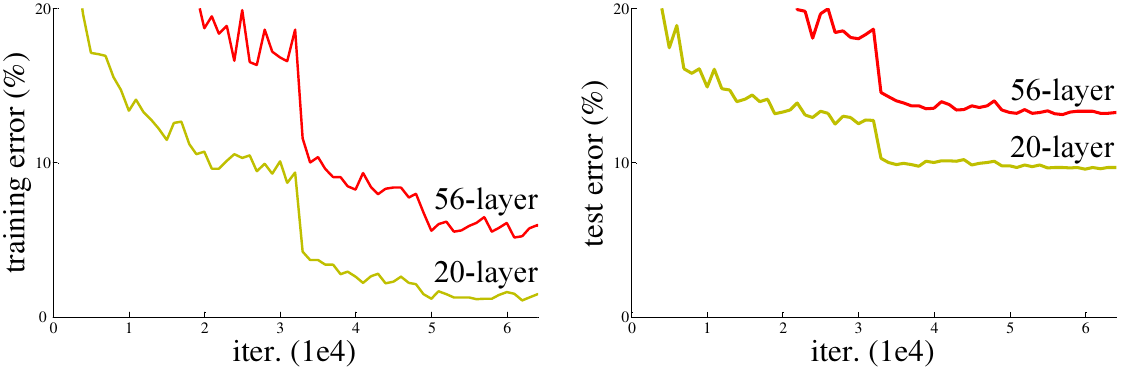}
\end{center}
\vspace{-1.2em}
\caption{Training error (left) and test error (right) on CIFAR-10 with 20-layer and 56-layer ``plain'' networks. The deeper network has higher training error, and thus test error. Similar phenomena on ImageNet is presented in Fig.~\ref{fig:imagenet}.}
\label{fig:teaser}
\vspace{-1em}
\end{figure}

When deeper networks are able to start converging, a \emph{degradation} problem has been exposed:  with the network depth increasing, accuracy gets saturated (which might be unsurprising) and then degrades rapidly. Unexpectedly, such degradation is \emph{not caused by overfitting}, and adding more layers to a suitably deep model leads to \emph{higher training error}, as reported in \cite{He2015a, Srivastava2015} and thoroughly verified by our experiments. Fig.~\ref{fig:teaser} shows a typical example.

The degradation (of training accuracy) indicates that not all systems are similarly easy to optimize. Let us consider a shallower architecture and its deeper counterpart that adds more layers onto it. There exists a solution \emph{by construction} to the deeper model: the added layers are \emph{identity} mapping, and the other layers are copied from the learned shallower model. The existence of this constructed solution indicates that a deeper model should produce no higher training error than its shallower counterpart. But experiments show that our current solvers on hand are unable to find solutions that are comparably good or better than the constructed solution (or unable to do so in feasible time).

In this paper, we address the degradation problem by introducing a \emph{deep residual learning} framework.
Instead of hoping each few stacked layers directly fit a desired underlying mapping, we explicitly let these layers fit a residual mapping. Formally, denoting the desired underlying mapping as $\mathcal{H}(\ve{x})$, we let the stacked nonlinear layers fit another mapping of $\mathcal{F}(\ve{x}):=\mathcal{H}(\ve{x})-\ve{x}$. The original mapping is recast into $\mathcal{F}(\ve{x})+\ve{x}$.
We hypothesize that it is easier to optimize the residual mapping than to optimize the original, unreferenced mapping. To the extreme, if an identity mapping were optimal, it would be easier to push the residual to zero than to fit an identity mapping by a stack of nonlinear layers.

\begin{figure}[t]
\centering
\hspace{48pt}
\includegraphics[width=0.9\linewidth]{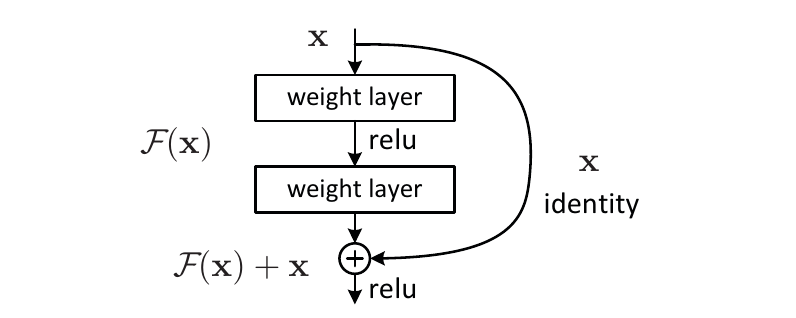}
\vspace{-.5em}
\caption{Residual learning: a building block.}
\label{fig:block}
\vspace{-1em}
\end{figure}

The formulation of $\mathcal{F}(\ve{x})+\ve{x}$ can be realized by feedforward neural networks with ``shortcut connections'' (Fig.~\ref{fig:block}). Shortcut connections \cite{Bishop1995,Ripley1996,Venables1999} are those skipping one or more layers. In our case, the shortcut connections simply perform \emph{identity} mapping, and their outputs are added to the outputs of the stacked layers (Fig.~\ref{fig:block}). Identity shortcut connections add neither extra parameter nor computational complexity. The entire network can still be trained end-to-end by SGD with backpropagation, and can be easily implemented using common libraries (\eg, Caffe \cite{Jia2014}) without modifying the solvers.

We present comprehensive experiments on ImageNet \cite{Russakovsky2014} to show the degradation problem and evaluate our method.
We show that: 1) Our extremely deep residual nets are easy to optimize, but the counterpart ``plain'' nets (that simply stack layers) exhibit higher training error when the depth increases; 2) Our deep residual nets can easily enjoy accuracy gains from greatly increased depth, producing results substantially better than previous networks.

Similar phenomena are also shown on the CIFAR-10 set \cite{Krizhevsky2009}, suggesting that the optimization difficulties and the effects of our method are not just akin to a particular dataset. We present successfully trained models on this dataset with over 100 layers, and explore models with over 1000 layers.

On the ImageNet classification dataset \cite{Russakovsky2014}, we obtain excellent results by extremely deep residual nets.
Our 152-layer residual net is the deepest network ever presented on ImageNet, while still having lower complexity than VGG nets \cite{Simonyan2015}. Our ensemble has \textbf{3.57\%} top-5 error on the ImageNet \emph{test} set, and \emph{won the 1st place in the ILSVRC 2015 classification competition}. The extremely deep representations also have excellent generalization performance on other recognition tasks, and lead us to further \emph{win the 1st places on: ImageNet detection, ImageNet localization, COCO detection, and COCO segmentation} in ILSVRC \& COCO 2015 competitions. This strong evidence shows that the residual learning principle is generic, and we expect that it is applicable in other vision and non-vision problems.

\section{Related Work}

\noindent\textbf{Residual Representations.}
In image recognition, VLAD \cite{Jegou2012} is a representation that encodes by the residual vectors with respect to a dictionary, and Fisher Vector \cite{Perronnin2007} can be formulated as a probabilistic version \cite{Jegou2012} of VLAD.
Both of them are powerful shallow representations for image retrieval and classification \cite{Chatfield2011,Vedaldi2008}.
For vector quantization, encoding residual vectors \cite{Jegou2011} is shown to be more effective than encoding original vectors.

In low-level vision and computer graphics, for solving Partial Differential Equations (PDEs), the widely used Multigrid method \cite{Briggs2000} reformulates the system as subproblems at multiple scales, where each subproblem is responsible for the residual solution between a coarser and a finer scale. An alternative to Multigrid is hierarchical basis preconditioning \cite{Szeliski1990,Szeliski2006}, which relies on variables that represent residual vectors between two scales. It has been shown \cite{Briggs2000,Szeliski1990,Szeliski2006} that these solvers converge much faster than standard solvers that are unaware of the residual nature of the solutions. These methods suggest that a good reformulation or preconditioning can simplify the optimization.

\vspace{6pt}
\noindent\textbf{Shortcut Connections.}
Practices and theories that lead to shortcut connections \cite{Bishop1995,Ripley1996,Venables1999} have been studied for a long time.
An early practice of training multi-layer perceptrons (MLPs) is to add a linear layer connected from the network input to the output \cite{Ripley1996,Venables1999}. In \cite{Szegedy2015,Lee2014}, a few intermediate layers are directly connected to auxiliary classifiers for addressing vanishing/exploding gradients. The papers of \cite{Schraudolph1998,Schraudolph1998a,Raiko2012,Vatanen2013} propose methods for centering layer responses, gradients, and propagated errors, implemented by shortcut connections. In \cite{Szegedy2015}, an ``inception'' layer is composed of a shortcut branch and a few deeper branches.

Concurrent with our work, ``highway networks'' \cite{Srivastava2015,Srivastava2015a} present shortcut connections with gating functions \cite{Hochreiter1997}. These gates are data-dependent and have parameters, in contrast to our identity shortcuts that are parameter-free. When a gated shortcut is ``closed'' (approaching zero), the layers in highway networks represent \emph{non-residual} functions. On the contrary, our formulation always learns residual functions; our identity shortcuts are never closed, and all information is always passed through, with additional residual functions to be learned. In addition, highway networks have not demonstrated accuracy gains with extremely increased depth (\eg, over 100 layers).

\section{Deep Residual Learning}

\subsection{Residual Learning}
\label{sec:motivation}

Let us consider $\mathcal{H}(\ve{x})$ as an underlying mapping to be fit by a few stacked layers (not necessarily the entire net), with $\ve{x}$ denoting the inputs to the first of these layers. If one hypothesizes that multiple nonlinear layers can asymptotically approximate complicated functions\footnote{This hypothesis, however, is still an open question. See \cite{Montufar2014}.}, then it is equivalent to hypothesize that they can asymptotically approximate the residual functions, \ie, $\mathcal{H}(\ve{x})-\ve{x}$ (assuming that the input and output are of the same dimensions).
So rather than expect stacked layers to approximate $\mathcal{H}(\ve{x})$, we explicitly let these layers approximate a residual function $\mathcal{F}(\ve{x}):=\mathcal{H}(\ve{x})-\ve{x}$. The original function thus becomes $\mathcal{F}(\ve{x})+\ve{x}$. Although both forms should be able to asymptotically approximate the desired functions (as hypothesized), the ease of learning might be different.

This reformulation is motivated by the counterintuitive phenomena about the degradation problem (Fig.~\ref{fig:teaser}, left). As we discussed in the introduction, if the added layers can be constructed as identity mappings, a deeper model should have training error no greater than its shallower counterpart. The degradation problem suggests that the solvers might have difficulties in approximating identity mappings by multiple nonlinear layers. With the residual learning reformulation, if identity mappings are optimal, the solvers may simply drive the weights of the multiple nonlinear layers toward zero to approach identity mappings.

In real cases, it is unlikely that identity mappings are optimal, but our reformulation may help to precondition the problem. If the optimal function is closer to an identity mapping than to a zero mapping, it should be easier for the solver to find the perturbations with reference to an identity mapping, than to learn the function as a new one. We show by experiments (Fig.~\ref{fig:std}) that the learned residual functions in general have small responses, suggesting that identity mappings provide reasonable preconditioning.

\subsection{Identity Mapping by Shortcuts}

We adopt residual learning to every few stacked layers.
A building block is shown in Fig.~\ref{fig:block}. Formally, in this paper we consider a building block defined as:
\begin{equation}\label{eq:identity}
\ve{y}= \mathcal{F}(\ve{x}, \{W_{i}\}) + \ve{x}.
\end{equation}
Here $\ve{x}$ and $\ve{y}$ are the input and output vectors of the layers considered. The function $\mathcal{F}(\ve{x}, \{W_{i}\})$ represents the residual mapping to be learned. For the example in Fig.~\ref{fig:block} that has two layers, $\mathcal{F}=W_{2}\sigma(W_{1}\ve{x})$ in which $\sigma$ denotes ReLU \cite{Nair2010} and the biases are omitted for simplifying notations. The operation $\mathcal{F}+\ve{x}$ is performed by a shortcut connection and element-wise addition. We adopt the second nonlinearity after the addition (\ie, $\sigma(\ve{y})$, see Fig.~\ref{fig:block}).

The shortcut connections in Eqn.(\ref{eq:identity}) introduce neither extra parameter nor computation complexity. This is not only attractive in practice but also important in our comparisons between plain and residual networks. We can fairly compare plain/residual networks that simultaneously have the same number of parameters, depth, width, and computational cost (except for the negligible element-wise addition).

The dimensions of $\ve{x}$ and $\mathcal{F}$ must be equal in Eqn.(\ref{eq:identity}). If this is not the case (\eg, when changing the input/output channels), we can perform a linear projection $W_{s}$ by the shortcut connections to match the dimensions:
\begin{equation}\label{eq:transform}
\ve{y}= \mathcal{F}(\ve{x}, \{W_{i}\}) + W_{s}\ve{x}.
\end{equation}
We can also use a square matrix $W_{s}$ in Eqn.(\ref{eq:identity}). But we will show by experiments that the identity mapping is sufficient for addressing the degradation problem and is economical, and thus $W_{s}$ is only used when matching dimensions.

The form of the residual function $\mathcal{F}$ is flexible. Experiments in this paper involve a function $\mathcal{F}$ that has two or three layers (Fig.~\ref{fig:block_deeper}), while more layers are possible. But if $\mathcal{F}$ has only a single layer, Eqn.(\ref{eq:identity}) is similar to a linear layer: $\ve{y}=W_1\ve{x}+\ve{x}$, for which we have not observed advantages.

We also note that although the above notations are about fully-connected layers for simplicity, they are applicable to convolutional layers. The function $\mathcal{F}(\ve{x}, \{W_{i}\})$ can represent multiple convolutional layers. The element-wise addition is performed on two feature maps, channel by channel.

\begin{figure}[t]
\begin{center}
\vspace{.5em}
\includegraphics[width=1.0\linewidth]{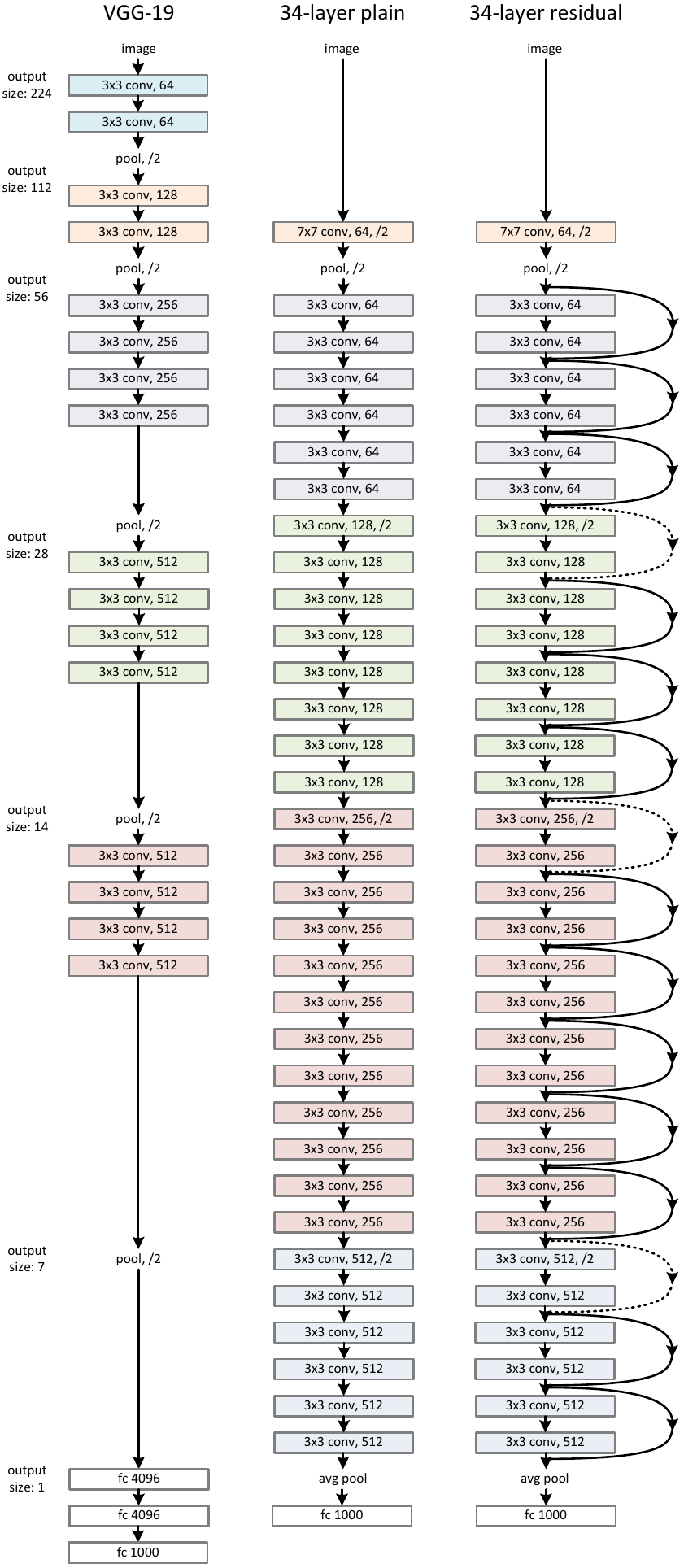}
\end{center}
\caption{Example network architectures for ImageNet. \textbf{Left}: the VGG-19 model \cite{Simonyan2015} (19.6 billion FLOPs) as a reference. \textbf{Middle}: a plain network with 34 parameter layers (3.6 billion FLOPs). \textbf{Right}: a residual network with 34 parameter layers (3.6 billion FLOPs). The dotted shortcuts increase dimensions. \textbf{Table~\ref{tab:arch}} shows more details and other variants.}
\label{fig:arch}
\vspace{-1em}
\end{figure}

\subsection{Network Architectures}

We have tested various plain/residual nets, and have observed consistent phenomena. To provide instances for discussion, we describe two models for ImageNet as follows.

\vspace{6pt}
\noindent\textbf{Plain Network.}
Our plain baselines (Fig.~\ref{fig:arch}, middle) are mainly inspired by the philosophy of VGG nets \cite{Simonyan2015} (Fig.~\ref{fig:arch}, left).
The convolutional layers mostly have 3$\times$3 filters and follow two simple design rules: (i) for the same output feature map size, the layers have the same number of filters; and (ii) if the feature map size is halved, the number of filters is doubled so as to preserve the time complexity per layer. We perform downsampling directly by convolutional layers that have a stride of 2.
The network ends with a global average pooling layer and a 1000-way fully-connected layer with softmax. The total number of weighted layers is 34 in Fig.~\ref{fig:arch} (middle).

It is worth noticing that our model has \emph{fewer} filters and \emph{lower} complexity than VGG nets \cite{Simonyan2015} (Fig.~\ref{fig:arch}, left). Our 34-layer baseline has 3.6 billion FLOPs (multiply-adds), which is only 18\% of VGG-19 (19.6 billion FLOPs).

\vspace{6pt}
\noindent\textbf{Residual Network.}
Based on the above plain network, we insert shortcut connections (Fig.~\ref{fig:arch}, right) which turn the network into its counterpart residual version.
The identity shortcuts (Eqn.(\ref{eq:identity})) can be directly used when the input and output are of the same dimensions (solid line shortcuts in Fig.~\ref{fig:arch}).
When the dimensions increase (dotted line shortcuts in Fig.~\ref{fig:arch}), we consider two options:
(A) The shortcut still performs identity mapping, with extra zero entries padded for increasing dimensions. This option introduces no extra parameter;
(B) The projection shortcut in Eqn.(\ref{eq:transform}) is used to match dimensions (done by 1$\times$1 convolutions).
For both options, when the shortcuts go across feature maps of two sizes, they are performed with a stride of 2.

\newcommand{\blocka}[2]{\multirow{3}{*}{\(\left[\begin{array}{c}\text{3$\times$3, #1}\\[-.1em] \text{3$\times$3, #1} \end{array}\right]\)$\times$#2}
}
\newcommand{\blockb}[3]{\multirow{3}{*}{\(\left[\begin{array}{c}\text{1$\times$1, #2}\\[-.1em] \text{3$\times$3, #2}\\[-.1em] \text{1$\times$1, #1}\end{array}\right]\)$\times$#3}
}
\renewcommand\arraystretch{1.1}
\setlength{\tabcolsep}{3pt}
\begin{table*}[t]
\begin{center}
\resizebox{0.7\linewidth}{!}{
\begin{tabular}{c|c|c|c|c|c|c}
\hline
layer name & output size & 18-layer & 34-layer & 50-layer & 101-layer & 152-layer \\
\hline
conv1 & 112$\times$112 & \multicolumn{5}{c}{7$\times$7, 64, stride 2}\\
\hline
\multirow{4}{*}{conv2\_x} & \multirow{4}{*}{56$\times$56} & \multicolumn{5}{c}{3$\times$3 max pool, stride 2} \\\cline{3-7}
  &  & \blocka{64}{2}  & \blocka{64}{3} & \blockb{256}{64}{3} & \blockb{256}{64}{3} & \blockb{256}{64}{3}\\
  &  &  &  &  &  &\\
  &  &  &  &  &  &\\
\hline
\multirow{3}{*}{conv3\_x} &  \multirow{3}{*}{28$\times$28}  & \blocka{128}{2}  & \blocka{128}{4}  & \blockb{512}{128}{4}  & \blockb{512}{128}{4}  &
                              \blockb{512}{128}{8}\\
  &  &  &  &  &  & \\
  &  &  &  &  &  & \\
\hline
\multirow{3}{*}{conv4\_x} & \multirow{3}{*}{14$\times$14}  & \blocka{256}{2}  & \blocka{256}{6}  & \blockb{1024}{256}{6}  & \blockb{1024}{256}{23} & \blockb{1024}{256}{36}\\
  &  &  &  &  & \\
  &  &  &  &  & \\
\hline
\multirow{3}{*}{conv5\_x} & \multirow{3}{*}{7$\times$7}  & \blocka{512}{2}  & \blocka{512}{3}  & \blockb{2048}{512}{3}  & \blockb{2048}{512}{3}
& \blockb{2048}{512}{3}\\
  &  &  &  &  &  & \\
  &  &  &  &  &  & \\
\hline
& 1$\times$1  & \multicolumn{5}{c}{average pool, 1000-d fc, softmax} \\
\hline
\multicolumn{2}{c|}{FLOPs} & 1.8$\times10^9$  & 3.6$\times10^9$  & 3.8$\times10^9$  & 7.6$\times10^9$  & 11.3$\times10^9$ \\
\hline
\end{tabular}
}
\end{center}
\vspace{-.5em}
\caption{Architectures for ImageNet. Building blocks are shown in brackets (see also Fig.~\ref{fig:block_deeper}), with the numbers of blocks stacked. Downsampling is performed by conv3\_1, conv4\_1, and conv5\_1 with a stride of 2.
}
\label{tab:arch}
\vspace{-.5em}
\end{table*}

\begin{figure*}[t]
\begin{center}
\includegraphics[width=0.86\linewidth]{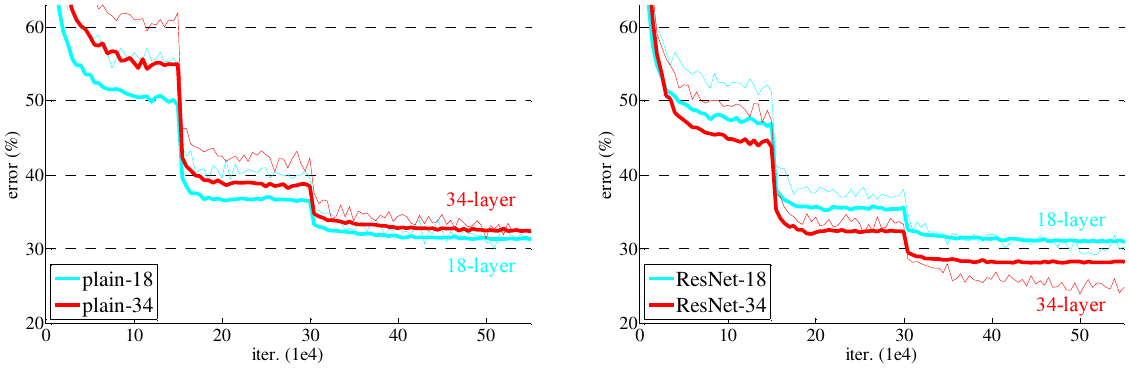}
\end{center}
\vspace{-1.2em}
\caption{Training on \textbf{ImageNet}. Thin curves denote training error, and bold curves denote validation error of the center crops. Left: plain networks of 18 and 34 layers. Right: ResNets of 18 and 34 layers. In this plot, the residual networks have no extra parameter compared to their plain counterparts.}
\label{fig:imagenet}
\end{figure*}

\subsection{Implementation}
\label{sec:impl}

Our implementation for ImageNet follows the practice in \cite{Krizhevsky2012,Simonyan2015}. The image is resized with its shorter side randomly sampled in $[256, 480]$ for scale augmentation \cite{Simonyan2015}. A 224$\times$224 crop is randomly sampled from an image or its horizontal flip, with the per-pixel mean subtracted \cite{Krizhevsky2012}. The standard color augmentation in \cite{Krizhevsky2012} is used.
We adopt batch normalization (BN) \cite{Ioffe2015} right after each convolution and before activation, following \cite{Ioffe2015}.
We initialize the weights as in \cite{He2015} and train all plain/residual nets from scratch.
We use SGD with a mini-batch size of 256. The learning rate starts from 0.1 and is divided by 10 when the error plateaus, and the models are trained for up to $60\times10^4$ iterations. We use a weight decay of 0.0001 and a momentum of 0.9. We do not use dropout \cite{Hinton2012}, following the practice in \cite{Ioffe2015}.

In testing, for comparison studies we adopt the standard 10-crop testing \cite{Krizhevsky2012}.
For best results, we adopt the fully-convolutional form as in \cite{Simonyan2015,He2015}, and average the scores at multiple scales (images are resized such that the shorter side is in $\{224, 256, 384, 480, 640\}$).

\section{Experiments}
\label{sec:exp}

\subsection{ImageNet Classification}
\label{sec:imagenet}

We evaluate our method on the ImageNet 2012 classification dataset \cite{Russakovsky2014} that consists of 1000 classes. The models are trained on the 1.28 million training images, and evaluated on the 50k validation images. We also obtain a final result on the 100k test images, reported by the test server. We evaluate both top-1 and top-5 error rates.

\vspace{6pt}
\noindent\textbf{Plain Networks.}
We first evaluate 18-layer and 34-layer plain nets. The 34-layer plain net is in Fig.~\ref{fig:arch} (middle). The 18-layer plain net is of a similar form. See Table~\ref{tab:arch} for detailed architectures.

The results in Table~\ref{tab:plain_vs_shortcut} show that the deeper 34-layer plain net has higher validation error than the shallower 18-layer plain net. To reveal the reasons, in Fig.~\ref{fig:imagenet} (left) we compare their training/validation errors during the training procedure. We have observed the degradation problem - the 34-layer plain net has higher \emph{training} error throughout the whole training procedure, even though the solution space of the 18-layer plain network is a subspace of that of the 34-layer one.

\newcolumntype{x}[1]{>{\centering}p{#1pt}}
\renewcommand\arraystretch{1.1}
\setlength{\tabcolsep}{8pt}
\begin{table}[t]
\begin{center}
\small
\begin{tabular}{l|x{42}|c}
\hline
            & plain & ResNet \\
\hline
18 layers & 27.94 & 27.88 \\
34 layers & 28.54  & \textbf{25.03}  \\
\hline
\end{tabular}
\end{center}
\vspace{-.5em}
\caption{Top-1 error (\%, 10-crop testing) on ImageNet validation. Here the ResNets have no extra parameter compared to their plain counterparts. Fig.~\ref{fig:imagenet} shows the training procedures.}
\label{tab:plain_vs_shortcut}
\end{table}

We argue that this optimization difficulty is \emph{unlikely} to be caused by vanishing gradients. These plain networks are trained with BN \cite{Ioffe2015}, which ensures forward propagated signals to have non-zero variances. We also verify that the backward propagated gradients exhibit healthy norms with BN. So neither forward nor backward signals vanish.
In fact, the 34-layer plain net is still able to achieve competitive accuracy (Table~\ref{tab:10crop}), suggesting that the solver works to some extent. We conjecture that the deep plain nets may have exponentially low convergence rates, which impact the reducing of the training error\footnote{We have experimented with more training iterations (3$\times$) and still observed the degradation problem, suggesting that this problem cannot be feasibly addressed by simply using more iterations.}.
The reason for such optimization difficulties will be studied in the future.

\vspace{6pt}
\noindent\textbf{Residual Networks.}
Next we evaluate 18-layer and 34-layer residual nets (\emph{ResNets}). The baseline architectures are the same as the above plain nets, expect that a shortcut connection is added to each pair of 3$\times$3 filters as in Fig.~\ref{fig:arch} (right). In the first comparison (Table~\ref{tab:plain_vs_shortcut} and Fig.~\ref{fig:imagenet} right), we use identity mapping for all shortcuts and zero-padding for increasing dimensions (option A). So they have \emph{no extra parameter} compared to the plain counterparts.

We have three major observations from Table~\ref{tab:plain_vs_shortcut} and Fig.~\ref{fig:imagenet}. First, the situation is reversed with residual learning -- the 34-layer ResNet is better than the 18-layer ResNet (by 2.8\%). More importantly, the 34-layer ResNet exhibits considerably lower training error and is generalizable to the validation data. This indicates that the degradation problem is well addressed in this setting and we manage to obtain accuracy gains from increased depth.

Second, compared to its plain counterpart, the 34-layer ResNet reduces the top-1 error by 3.5\% (Table~\ref{tab:plain_vs_shortcut}), resulting from the successfully reduced training error (Fig.~\ref{fig:imagenet} right \vs left). This comparison verifies the effectiveness of residual learning on extremely deep systems.

Last, we also note that the 18-layer plain/residual nets are comparably accurate (Table~\ref{tab:plain_vs_shortcut}), but the 18-layer ResNet converges faster (Fig.~\ref{fig:imagenet} right \vs left).
When the net is ``not overly deep'' (18 layers here), the current SGD solver is still able to find good solutions to the plain net. In this case, the ResNet eases the optimization by providing faster convergence at the early stage.

\begin{table}[t]
\setlength{\tabcolsep}{8pt}
\begin{center}
\small
\begin{tabular}{l|cc}
\hline
  \footnotesize model          & \footnotesize top-1 err. & \footnotesize top-5 err. \\
\hline
\footnotesize VGG-16 \cite{Simonyan2015} & 28.07 & 9.33\\
\footnotesize GoogLeNet \cite{Szegedy2015} & - & 9.15 \\
\footnotesize PReLU-net \cite{He2015}  & 24.27 & 7.38 \\
\hline
\hline
\footnotesize plain-34 & 28.54 & 10.02 \\
\footnotesize ResNet-34 A & 25.03 & 7.76 \\
\footnotesize ResNet-34 B & 24.52 & 7.46 \\
\footnotesize ResNet-34 C & 24.19 & 7.40 \\
\hline
\footnotesize ResNet-50 & 22.85 & 6.71 \\
\footnotesize ResNet-101 & 21.75 & 6.05 \\
\footnotesize ResNet-152 & \textbf{21.43} & \textbf{5.71} \\
\hline
\end{tabular}
\end{center}
\vspace{-.5em}
\caption{Error rates (\%, \textbf{10-crop} testing) on ImageNet validation.
VGG-16 is based on our test. ResNet-50/101/152 are of option B that only uses projections for increasing dimensions.}
\label{tab:10crop}
\vspace{-.5em}
\end{table}

\begin{table}[t]
\setlength{\tabcolsep}{8pt}
\small
\begin{center}
\begin{tabular}{l|c c}
\hline
\footnotesize method & \footnotesize top-1 err. & \footnotesize top-5 err.\\
\hline
VGG \cite{Simonyan2015} (ILSVRC'14) & - & 8.43$^{\dag}$\\
GoogLeNet \cite{Szegedy2015} (ILSVRC'14) & - & 7.89\\
\hline
VGG \cite{Simonyan2015} \footnotesize (v5) & 24.4 & 7.1\\
PReLU-net \cite{He2015} & 21.59 & 5.71 \\
BN-inception \cite{Ioffe2015} & 21.99 & 5.81 \\\hline
ResNet-34 B & 21.84 & 5.71 \\
ResNet-34 C & 21.53 & 5.60 \\
ResNet-50 & 20.74 & 5.25 \\
ResNet-101 & 19.87 & 4.60 \\
ResNet-152 & \textbf{19.38} & \textbf{4.49} \\
\hline
\end{tabular}
\end{center}
\vspace{-.5em}
\caption{Error rates (\%) of \textbf{single-model} results on the ImageNet validation set (except $^{\dag}$ reported on the test set).}
\label{tab:single}
\setlength{\tabcolsep}{12pt}
\small
\begin{center}
\begin{tabular}{l|c}
\hline
\footnotesize method & top-5 err. (\textbf{test}) \\
\hline
VGG \cite{Simonyan2015} (ILSVRC'14) & 7.32\\
GoogLeNet \cite{Szegedy2015} (ILSVRC'14) & 6.66\\
\hline
VGG \cite{Simonyan2015} \footnotesize (v5) & 6.8 \\
PReLU-net \cite{He2015} & 4.94 \\
BN-inception \cite{Ioffe2015} & 4.82 \\\hline
\textbf{ResNet (ILSVRC'15)} & \textbf{3.57} \\
\hline
\end{tabular}
\end{center}
\vspace{-.5em}
\caption{Error rates (\%) of \textbf{ensembles}. The top-5 error is on the test set of ImageNet and reported by the test server.}
\label{tab:ensemble}
\end{table}

\vspace{6pt}
\noindent\textbf{Identity \vs Projection Shortcuts.}
We have shown that parameter-free, identity shortcuts help with training. Next we investigate projection shortcuts (Eqn.(\ref{eq:transform})).
In Table~\ref{tab:10crop} we compare three options: (A) zero-padding shortcuts are used for increasing dimensions, and all shortcuts are parameter-free (the same as Table~\ref{tab:plain_vs_shortcut} and Fig.~\ref{fig:imagenet} right); (B) projection shortcuts are used for increasing dimensions, and other shortcuts are identity; and (C) all shortcuts are projections.

Table~\ref{tab:10crop} shows that all three options are considerably better than the plain counterpart.
B is slightly better than A. We argue that this is because the zero-padded dimensions in A indeed have no residual learning. C is marginally better than B, and we attribute this to the extra parameters introduced by many (thirteen) projection shortcuts. But the small differences among A/B/C indicate that projection shortcuts are not essential for addressing the degradation problem. So we do not use option C in the rest of this paper, to reduce memory/time complexity and model sizes. Identity shortcuts are particularly important for not increasing the complexity of the bottleneck architectures that are introduced below.

\begin{figure}[t]
\begin{center}
\hspace{12pt}
\includegraphics[width=0.85\linewidth]{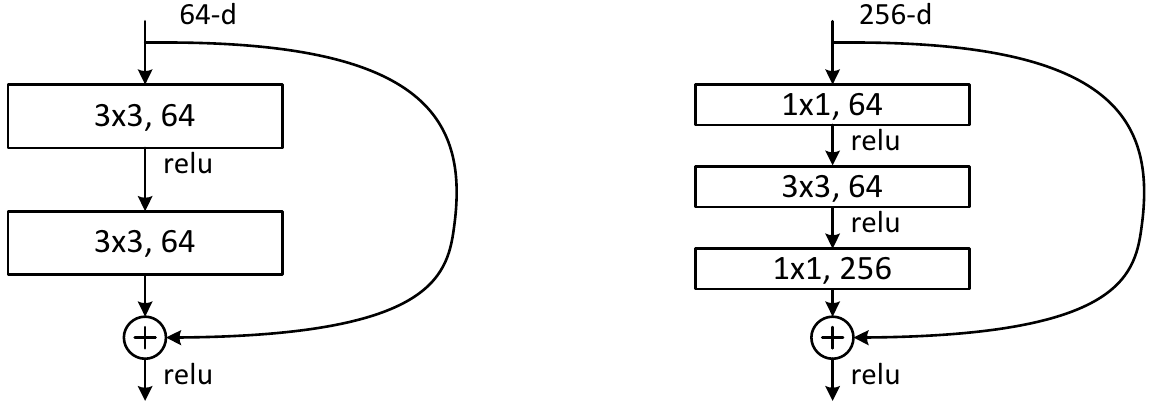}
\end{center}
\caption{A deeper residual function $\mathcal{F}$ for ImageNet. Left: a building block (on 56$\times$56 feature maps) as in Fig.~\ref{fig:arch} for ResNet-34. Right: a ``bottleneck'' building block for ResNet-50/101/152.}
\label{fig:block_deeper}
\vspace{-.6em}
\end{figure}

\vspace{6pt}
\noindent\textbf{Deeper Bottleneck Architectures.} Next we describe our deeper nets for ImageNet. Because of concerns on the training time that we can afford, we modify the building block as a \emph{bottleneck} design\footnote{Deeper \emph{non}-bottleneck ResNets (\eg, Fig.~\ref{fig:block_deeper} left) also gain accuracy from increased depth (as shown on CIFAR-10), but are not as economical as the bottleneck ResNets. So the usage of bottleneck designs is mainly due to practical considerations. We further note that the degradation problem of plain nets is also witnessed for the bottleneck designs.}.
For each residual function $\mathcal{F}$, we use a stack of 3 layers instead of 2 (Fig.~\ref{fig:block_deeper}). The three layers are 1$\times$1, 3$\times$3, and 1$\times$1 convolutions, where the 1$\times$1 layers are responsible for reducing and then increasing (restoring) dimensions, leaving the 3$\times$3 layer a bottleneck with smaller input/output dimensions.
Fig.~\ref{fig:block_deeper} shows an example, where both designs have similar time complexity.

The parameter-free identity shortcuts are particularly important for the bottleneck architectures. If the identity shortcut in Fig.~\ref{fig:block_deeper} (right) is replaced with projection, one can show that the time complexity and model size are doubled, as the shortcut is connected to the two high-dimensional ends. So identity shortcuts lead to more efficient models for the bottleneck designs.

\textbf{50-layer ResNet:} We replace each 2-layer block in the 34-layer net with this 3-layer bottleneck block, resulting in a 50-layer ResNet (Table~\ref{tab:arch}). We use option B for increasing dimensions.
This model has 3.8 billion FLOPs.

\textbf{101-layer and 152-layer ResNets:} We construct 101-layer and 152-layer ResNets by using more 3-layer blocks (Table~\ref{tab:arch}).
Remarkably, although the depth is significantly increased, the 152-layer ResNet (11.3 billion FLOPs) still has \emph{lower complexity} than VGG-16/19 nets (15.3/19.6 billion FLOPs).

The 50/101/152-layer ResNets are more accurate than the 34-layer ones by considerable margins (Table~\ref{tab:10crop} and~\ref{tab:single}). We do not observe the degradation problem and thus enjoy significant accuracy gains from considerably increased depth. The benefits of depth are witnessed for all evaluation metrics (Table~\ref{tab:10crop} and~\ref{tab:single}).

\begin{figure*}[t]
\begin{center}
\includegraphics[width=0.8\linewidth]{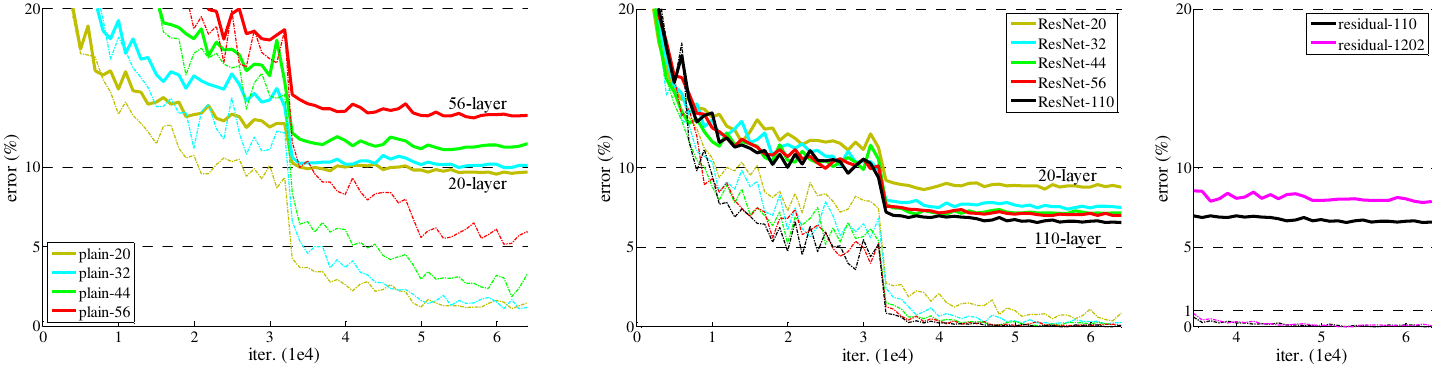}
\end{center}
\vspace{-1.5em}
\caption{Training on \textbf{CIFAR-10}. Dashed lines denote training error, and bold lines denote testing error. \textbf{Left}: plain networks. The error of plain-110 is higher than 60\% and not displayed. \textbf{Middle}: ResNets. \textbf{Right}: ResNets with 110 and 1202 layers.}
\label{fig:cifar}
\end{figure*}

\vspace{6pt}
\noindent\textbf{Comparisons with State-of-the-art Methods.}
In Table~\ref{tab:single} we compare with the previous best single-model results.
Our baseline 34-layer ResNets have achieved very competitive accuracy.
Our 152-layer ResNet has a single-model top-5 validation error of 4.49\%. This single-model result outperforms all previous ensemble results (Table~\ref{tab:ensemble}).
We combine six models of different depth to form an ensemble (only with two 152-layer ones at the time of submitting). This leads to \textbf{3.57\%} top-5 error on the test set (Table~\ref{tab:ensemble}). \emph{This entry won the 1st place in ILSVRC 2015.}

\subsection{CIFAR-10 and Analysis}

We conducted more studies on the CIFAR-10 dataset \cite{Krizhevsky2009}, which consists of 50k training images and 10k testing images in 10 classes. We present experiments trained on the training set and evaluated on the test set.
Our focus is on the behaviors of extremely deep networks, but not on pushing the state-of-the-art results, so we intentionally use simple architectures as follows.

The plain/residual architectures follow the form in Fig.~\ref{fig:arch} (middle/right).
The network inputs are 32$\times$32 images, with the per-pixel mean subtracted. The first layer is 3$\times$3 convolutions. Then we use a stack of $6n$ layers with 3$\times$3 convolutions on the feature maps of sizes $\{32, 16, 8\}$ respectively, with 2$n$ layers for each feature map size. The numbers of filters are $\{16, 32, 64\}$ respectively. The subsampling is performed by convolutions with a stride of 2. The network ends with a global average pooling, a 10-way fully-connected layer, and softmax. There are totally 6$n$+2 stacked weighted layers. The following table summarizes the architecture:
\renewcommand\arraystretch{1.1}
\begin{center}
\small
\setlength{\tabcolsep}{8pt}
\begin{tabular}{c|c|c|c}
\hline
output map size & 32$\times$32 & 16$\times$16 & 8$\times$8 \\
\hline
\# layers & 1+2$n$ & 2$n$ & 2$n$\\
\# filters & 16 & 32 & 64\\
\hline
\end{tabular}
\end{center}
When shortcut connections are used, they are connected to the pairs of 3$\times$3 layers (totally $3n$ shortcuts). On this dataset we use identity shortcuts in all cases (\ie, option A), so our residual models have exactly the same depth, width, and number of parameters as the plain counterparts.

\renewcommand\arraystretch{1.05}
\setlength{\tabcolsep}{5pt}
\begin{table}[t]
\begin{center}
\small
\resizebox{1.0\linewidth}{!}{
\begin{tabular}{c|c|c|l}
\hline
  \multicolumn{3}{c|}{method}   & error (\%) \\
\hline
\multicolumn{3}{c|}{Maxout \cite{Goodfellow2013}} & 9.38 \\
\multicolumn{3}{c|}{NIN \cite{Lin2013}} & 8.81 \\
\multicolumn{3}{c|}{DSN \cite{Lee2014}} & 8.22 \\
\hline
 & \# layers & \# params &  \\
\hline
FitNet \cite{Romero2015} & 19 & 2.5M  & 8.39 \\
Highway \cite{Srivastava2015,Srivastava2015a} & 19 & 2.3M  & 7.54 \footnotesize (7.72$\pm$0.16) \\
Highway \cite{Srivastava2015,Srivastava2015a} & 32 & 1.25M  & 8.80 \\
\hline
ResNet & 20 & 0.27M & 8.75 \\
ResNet & 32 & 0.46M & 7.51 \\
ResNet & 44 & 0.66M & 7.17 \\
ResNet & 56 & 0.85M & 6.97 \\
ResNet & 110 & 1.7M & \textbf{6.43} \footnotesize (6.61$\pm$0.16) \\ 
ResNet & 1202 & 19.4M & 7.93 \\
\hline
\end{tabular}
}
\end{center}
\caption{Classification error on the \textbf{CIFAR-10} test set. All methods are with data augmentation. For ResNet-110, we run it 5 times and show ``best (mean$\pm$std)'' as in \cite{Srivastava2015a}.
}
\label{tab:cifar}
\end{table}

We use a weight decay of 0.0001 and momentum of 0.9, and adopt the weight initialization in \cite{He2015} and BN \cite{Ioffe2015} but with no dropout. These models are trained with a mini-batch size of 128 on two GPUs. We start with a learning rate of 0.1, divide it by 10 at 32k and 48k iterations, and terminate training at 64k iterations, which is determined on a 45k/5k train/val split. We follow the simple data augmentation in \cite{Lee2014} for training: 4 pixels are padded on each side, and a 32$\times$32 crop is randomly sampled from the padded image or its horizontal flip. For testing, we only evaluate the single view of the original 32$\times$32 image.

We compare $n=\{3,5,7,9\}$, leading to 20, 32, 44, and 56-layer networks.
Fig.~\ref{fig:cifar} (left) shows the behaviors of the plain nets. The deep plain nets suffer from increased depth, and exhibit higher training error when going deeper. This phenomenon is similar to that on ImageNet (Fig.~\ref{fig:imagenet}, left) and on MNIST (see \cite{Srivastava2015}), suggesting that such an optimization difficulty is a fundamental problem.

Fig.~\ref{fig:cifar} (middle) shows the behaviors of ResNets. Also similar to the ImageNet cases (Fig.~\ref{fig:imagenet}, right), our ResNets manage to overcome the optimization difficulty and demonstrate accuracy gains when the depth increases.

We further explore $n=18$ that leads to a 110-layer ResNet. In this case, we find that the initial learning rate of 0.1 is slightly too large to start converging\footnote{With an initial learning rate of 0.1, it starts converging ($<$90\% error) after several epochs, but still reaches similar accuracy.}. So we use 0.01 to warm up the training until the training error is below 80\% (about 400 iterations), and then go back to 0.1 and continue training. The rest of the learning schedule is as done previously. This 110-layer network converges well (Fig.~\ref{fig:cifar}, middle). It has \emph{fewer} parameters than other deep and thin networks such as FitNet \cite{Romero2015} and Highway \cite{Srivastava2015} (Table~\ref{tab:cifar}), yet is among the state-of-the-art results (6.43\%, Table~\ref{tab:cifar}).

\begin{figure}[t]
\begin{center}
\includegraphics[width=0.9\linewidth]{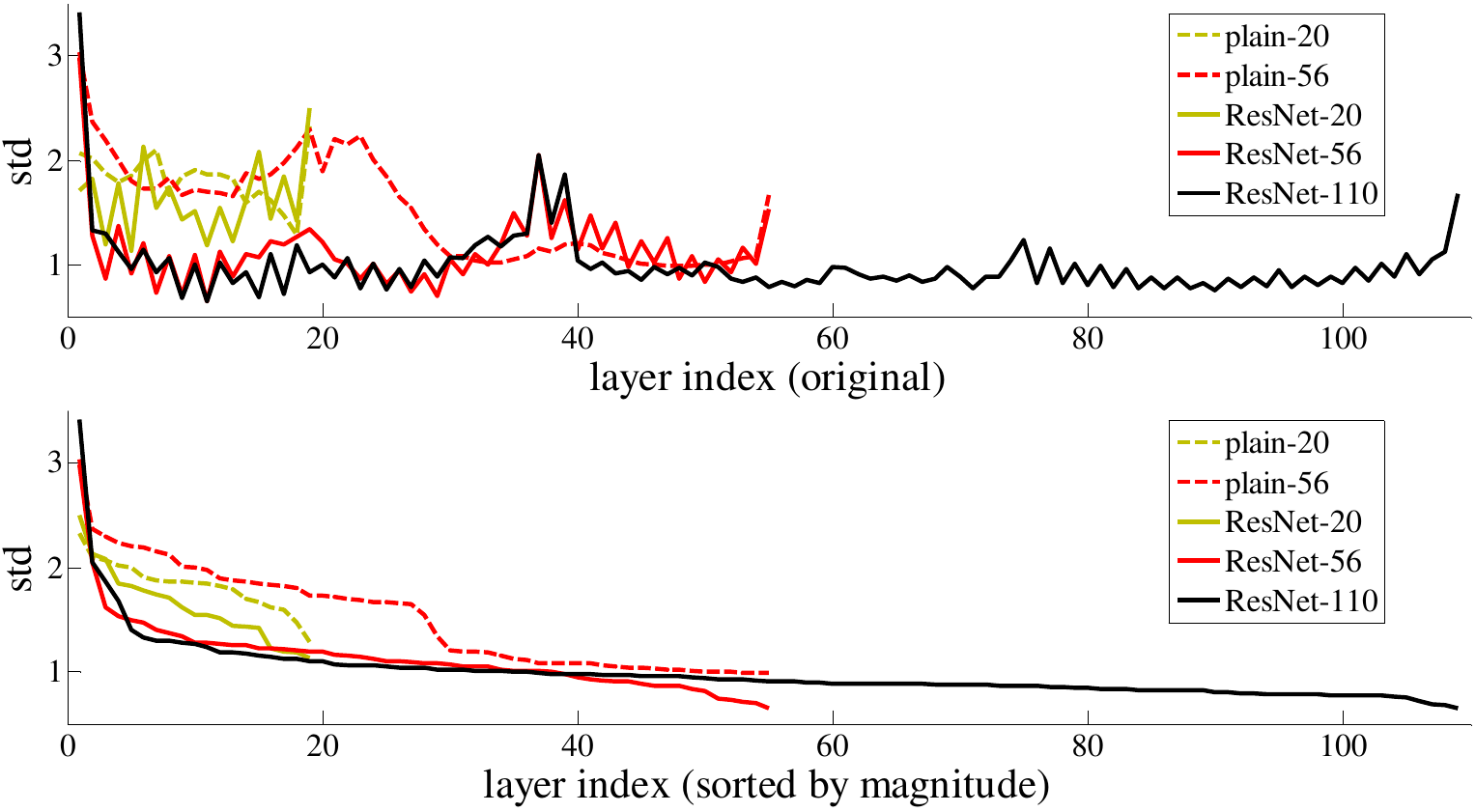}
\end{center}
\vspace{-1.5em}
\caption{Standard deviations (std) of layer responses on CIFAR-10. The responses are the outputs of each 3$\times$3 layer, after BN and before nonlinearity. \textbf{Top}: the layers are shown in their original order. \textbf{Bottom}: the responses are ranked in descending order.}
\label{fig:std}
\end{figure}

\vspace{6pt}
\noindent\textbf{Analysis of Layer Responses.}
Fig.~\ref{fig:std} shows the standard deviations (std) of the layer responses. The responses are the outputs of each 3$\times$3 layer, after BN and before other nonlinearity (ReLU/addition). For ResNets, this analysis reveals the response strength of the residual functions.
Fig.~\ref{fig:std} shows that ResNets have generally smaller responses than their plain counterparts. These results support our basic motivation (Sec.\ref{sec:motivation}) that the residual functions might be generally closer to zero than the non-residual functions.
We also notice that the deeper ResNet has smaller magnitudes of responses, as evidenced by the comparisons among ResNet-20, 56, and 110 in Fig.~\ref{fig:std}. When there are more layers, an individual layer of ResNets tends to modify the signal less.

\vspace{6pt}
\noindent\textbf{Exploring Over 1000 layers.}
We explore an aggressively deep model of over 1000 layers. We set $n=200$ that leads to a 1202-layer network, which is trained as described above. Our method shows \emph{no optimization difficulty}, and this $10^3$-layer network is able to achieve \emph{training error} $<$0.1\% (Fig.~\ref{fig:cifar}, right). Its test error is still fairly good (7.93\%, Table~\ref{tab:cifar}).

But there are still open problems on such aggressively deep models.
The testing result of this 1202-layer network is worse than that of our 110-layer network, although both have similar training error. We argue that this is because of overfitting.
The 1202-layer network may be unnecessarily large (19.4M) for this small dataset. Strong regularization such as maxout \cite{Goodfellow2013} or dropout \cite{Hinton2012} is applied to obtain the best results (\cite{Goodfellow2013,Lin2013,Lee2014,Romero2015}) on this dataset.
In this paper, we use no maxout/dropout and just simply impose regularization via deep and thin architectures by design, without distracting from the focus on the difficulties of optimization. But combining with stronger regularization may improve results, which we will study in the future.

\subsection{Object Detection on PASCAL and MS COCO}

\renewcommand\arraystretch{1.05}
\setlength{\tabcolsep}{8pt}
\begin{table}[t]
\begin{center}
\small
\begin{tabular}{c|c|c}
\hline
training data & 07+12 & 07++12 \\
\hline
test data & VOC 07 test & VOC 12 test \\
\hline
VGG-16 & 73.2 & 70.4 \\
ResNet-101 & \textbf{76.4} & \textbf{73.8} \\
\hline
\end{tabular}
\end{center}
\vspace{-.5em}
\caption{Object detection mAP (\%) on the PASCAL VOC 2007/2012 test sets using \textbf{baseline} Faster R-CNN. See also Table~\ref{tab:voc07_all} and \ref{tab:voc12_all} for better results.
}
\vspace{-.5em}
\label{tab:detection_voc}
\setlength{\tabcolsep}{5pt}
\begin{center}
\small
\begin{tabular}{c|c|c}
\hline
metric & ~~~mAP@.5~~~ & mAP@[.5, .95] \\
\hline
VGG-16 & 41.5 & 21.2 \\
ResNet-101 & \textbf{48.4} & \textbf{27.2} \\
\hline
\end{tabular}
\end{center}
\vspace{-.5em}
\caption{Object detection mAP (\%) on the COCO validation set using \textbf{baseline} Faster R-CNN. See also Table~\ref{tab:detection_coco_improve} for better results.
}
\vspace{-.5em}
\label{tab:detection_coco}
\end{table}

Our method has good generalization performance on other recognition tasks. Table~\ref{tab:detection_voc} and ~\ref{tab:detection_coco} show the object detection baseline results on PASCAL VOC 2007 and 2012 \cite{Everingham2010} and COCO \cite{Lin2014}. We adopt \emph{Faster R-CNN} \cite{Ren2015} as the detection method. Here we are interested in the improvements of replacing VGG-16 \cite{Simonyan2015} with ResNet-101. The detection implementation (see appendix) of using both models is the same, so the gains can only be attributed to better networks. Most remarkably, on the challenging COCO dataset we obtain a 6.0\% increase in COCO's standard metric (mAP@[.5, .95]), which is a 28\% relative improvement. This gain is solely due to the learned representations.

Based on deep residual nets, we won the 1st places in several tracks in ILSVRC \& COCO 2015 competitions: ImageNet detection, ImageNet localization, COCO detection, and COCO segmentation. The details are in the appendix.


{
\footnotesize
\bibliographystyle{ieee}
\bibliography{residual_v1_arxiv_release}

\begin{thebibliography}{10}\itemsep=-1pt

\bibitem{Bengio1994}
Y.~Bengio, P.~Simard, and P.~Frasconi.
\newblock Learning long-term dependencies with gradient descent is difficult.
\newblock {\em IEEE Transactions on Neural Networks}, 5(2):157--166, 1994.

\bibitem{Bishop1995}
C.~M. Bishop.
\newblock {\em Neural networks for pattern recognition}.
\newblock Oxford university press, 1995.

\bibitem{Briggs2000}
W.~L. Briggs, S.~F. McCormick, et~al.
\newblock {\em {A Multigrid Tutorial}}.
\newblock Siam, 2000.

\bibitem{Chatfield2011}
K.~Chatfield, V.~Lempitsky, A.~Vedaldi, and A.~Zisserman.
\newblock The devil is in the details: an evaluation of recent feature encoding
  methods.
\newblock In {\em BMVC}, 2011.

\bibitem{Everingham2010}
M.~Everingham, L.~Van~Gool, C.~K. Williams, J.~Winn, and A.~Zisserman.
\newblock {The Pascal Visual Object Classes (VOC) Challenge}.
\newblock {\em IJCV}, pages 303--338, 2010.

\bibitem{Gidaris2015}
S.~Gidaris and N.~Komodakis.
\newblock Object detection via a multi-region \& semantic segmentation-aware
  cnn model.
\newblock In {\em ICCV}, 2015.

\bibitem{Girshick2015}
R.~Girshick.
\newblock {Fast R-CNN}.
\newblock In {\em ICCV}, 2015.

\bibitem{Girshick2014}
R.~Girshick, J.~Donahue, T.~Darrell, and J.~Malik.
\newblock Rich feature hierarchies for accurate object detection and semantic
  segmentation.
\newblock In {\em CVPR}, 2014.

\bibitem{Glorot2010}
X.~Glorot and Y.~Bengio.
\newblock Understanding the difficulty of training deep feedforward neural
  networks.
\newblock In {\em AISTATS}, 2010.

\bibitem{Goodfellow2013}
I.~J. Goodfellow, D.~Warde-Farley, M.~Mirza, A.~Courville, and Y.~Bengio.
\newblock Maxout networks.
\newblock {\em arXiv:1302.4389}, 2013.

\bibitem{He2015a}
K.~He and J.~Sun.
\newblock Convolutional neural networks at constrained time cost.
\newblock In {\em CVPR}, 2015.

\bibitem{He2014}
K.~He, X.~Zhang, S.~Ren, and J.~Sun.
\newblock Spatial pyramid pooling in deep convolutional networks for visual
  recognition.
\newblock In {\em ECCV}, 2014.

\bibitem{He2015}
K.~He, X.~Zhang, S.~Ren, and J.~Sun.
\newblock Delving deep into rectifiers: Surpassing human-level performance on
  imagenet classification.
\newblock In {\em ICCV}, 2015.

\bibitem{Hinton2012}
G.~E. Hinton, N.~Srivastava, A.~Krizhevsky, I.~Sutskever, and R.~R.
  Salakhutdinov.
\newblock Improving neural networks by preventing co-adaptation of feature
  detectors.
\newblock {\em arXiv:1207.0580}, 2012.

\bibitem{Hochreiter1997}
S.~Hochreiter and J.~Schmidhuber.
\newblock Long short-term memory.
\newblock {\em Neural computation}, 9(8):1735--1780, 1997.

\bibitem{Ioffe2015}
S.~Ioffe and C.~Szegedy.
\newblock Batch normalization: Accelerating deep network training by reducing
  internal covariate shift.
\newblock In {\em ICML}, 2015.

\bibitem{Jegou2011}
H.~Jegou, M.~Douze, and C.~Schmid.
\newblock Product quantization for nearest neighbor search.
\newblock {\em TPAMI}, 33, 2011.

\bibitem{Jegou2012}
H.~Jegou, F.~Perronnin, M.~Douze, J.~Sanchez, P.~Perez, and C.~Schmid.
\newblock Aggregating local image descriptors into compact codes.
\newblock {\em TPAMI}, 2012.

\bibitem{Jia2014}
Y.~Jia, E.~Shelhamer, J.~Donahue, S.~Karayev, J.~Long, R.~Girshick,
  S.~Guadarrama, and T.~Darrell.
\newblock Caffe: Convolutional architecture for fast feature embedding.
\newblock {\em arXiv:1408.5093}, 2014.

\bibitem{Krizhevsky2009}
A.~Krizhevsky.
\newblock Learning multiple layers of features from tiny images.
\newblock {\em Tech Report}, 2009.

\bibitem{Krizhevsky2012}
A.~Krizhevsky, I.~Sutskever, and G.~Hinton.
\newblock Imagenet classification with deep convolutional neural networks.
\newblock In {\em NIPS}, 2012.

\bibitem{LeCun1989}
Y.~LeCun, B.~Boser, J.~S. Denker, D.~Henderson, R.~E. Howard, W.~Hubbard, and
  L.~D. Jackel.
\newblock Backpropagation applied to handwritten zip code recognition.
\newblock {\em Neural computation}, 1989.

\bibitem{LeCun1998}
Y.~LeCun, L.~Bottou, G.~B. Orr, and K.-R. M{\"u}ller.
\newblock Efficient backprop.
\newblock In {\em Neural Networks: Tricks of the Trade}, pages 9--50. Springer,
  1998.

\bibitem{Lee2014}
C.-Y. Lee, S.~Xie, P.~Gallagher, Z.~Zhang, and Z.~Tu.
\newblock Deeply-supervised nets.
\newblock {\em arXiv:1409.5185}, 2014.

\bibitem{Lin2013}
M.~Lin, Q.~Chen, and S.~Yan.
\newblock Network in network.
\newblock {\em arXiv:1312.4400}, 2013.

\bibitem{Lin2014}
T.-Y. Lin, M.~Maire, S.~Belongie, J.~Hays, P.~Perona, D.~Ramanan,
  P.~Doll{\'a}r, and C.~L. Zitnick.
\newblock {Microsoft COCO: Common objects in context}.
\newblock In {\em ECCV}. 2014.

\bibitem{Long2015}
J.~Long, E.~Shelhamer, and T.~Darrell.
\newblock Fully convolutional networks for semantic segmentation.
\newblock In {\em CVPR}, 2015.

\bibitem{Montufar2014}
G.~Mont{\'u}far, R.~Pascanu, K.~Cho, and Y.~Bengio.
\newblock On the number of linear regions of deep neural networks.
\newblock In {\em NIPS}, 2014.

\bibitem{Nair2010}
V.~Nair and G.~E. Hinton.
\newblock Rectified linear units improve restricted boltzmann machines.
\newblock In {\em ICML}, 2010.

\bibitem{Perronnin2007}
F.~Perronnin and C.~Dance.
\newblock Fisher kernels on visual vocabularies for image categorization.
\newblock In {\em CVPR}, 2007.

\bibitem{Raiko2012}
T.~Raiko, H.~Valpola, and Y.~LeCun.
\newblock Deep learning made easier by linear transformations in perceptrons.
\newblock In {\em AISTATS}, 2012.

\bibitem{Ren2015}
S.~Ren, K.~He, R.~Girshick, and J.~Sun.
\newblock {Faster R-CNN}: Towards real-time object detection with region
  proposal networks.
\newblock In {\em NIPS}, 2015.

\bibitem{Ren2015a}
S.~Ren, K.~He, R.~Girshick, X.~Zhang, and J.~Sun.
\newblock Object detection networks on convolutional feature maps.
\newblock {\em arXiv:1504.06066}, 2015.

\bibitem{Ripley1996}
B.~D. Ripley.
\newblock {\em Pattern recognition and neural networks}.
\newblock Cambridge university press, 1996.

\bibitem{Romero2015}
A.~Romero, N.~Ballas, S.~E. Kahou, A.~Chassang, C.~Gatta, and Y.~Bengio.
\newblock Fitnets: Hints for thin deep nets.
\newblock In {\em ICLR}, 2015.

\bibitem{Russakovsky2014}
O.~Russakovsky, J.~Deng, H.~Su, J.~Krause, S.~Satheesh, S.~Ma, Z.~Huang,
  A.~Karpathy, A.~Khosla, M.~Bernstein, et~al.
\newblock Imagenet large scale visual recognition challenge.
\newblock {\em arXiv:1409.0575}, 2014.

\bibitem{Saxe2013}
A.~M. Saxe, J.~L. McClelland, and S.~Ganguli.
\newblock Exact solutions to the nonlinear dynamics of learning in deep linear
  neural networks.
\newblock {\em arXiv:1312.6120}, 2013.

\bibitem{Schraudolph1998a}
N.~N. Schraudolph.
\newblock Accelerated gradient descent by factor-centering decomposition.
\newblock Technical report, 1998.

\bibitem{Schraudolph1998}
N.~N. Schraudolph.
\newblock Centering neural network gradient factors.
\newblock In {\em Neural Networks: Tricks of the Trade}, pages 207--226.
  Springer, 1998.

\bibitem{Sermanet2014}
P.~Sermanet, D.~Eigen, X.~Zhang, M.~Mathieu, R.~Fergus, and Y.~LeCun.
\newblock Overfeat: Integrated recognition, localization and detection using
  convolutional networks.
\newblock In {\em ICLR}, 2014.

\bibitem{Simonyan2015}
K.~Simonyan and A.~Zisserman.
\newblock Very deep convolutional networks for large-scale image recognition.
\newblock In {\em ICLR}, 2015.

\bibitem{Srivastava2015}
R.~K. Srivastava, K.~Greff, and J.~Schmidhuber.
\newblock Highway networks.
\newblock {\em arXiv:1505.00387}, 2015.

\bibitem{Srivastava2015a}
R.~K. Srivastava, K.~Greff, and J.~Schmidhuber.
\newblock Training very deep networks.
\newblock {\em 1507.06228}, 2015.

\bibitem{Szegedy2015}
C.~Szegedy, W.~Liu, Y.~Jia, P.~Sermanet, S.~Reed, D.~Anguelov, D.~Erhan,
  V.~Vanhoucke, and A.~Rabinovich.
\newblock Going deeper with convolutions.
\newblock In {\em CVPR}, 2015.

\bibitem{Szeliski1990}
R.~Szeliski.
\newblock Fast surface interpolation using hierarchical basis functions.
\newblock {\em TPAMI}, 1990.

\bibitem{Szeliski2006}
R.~Szeliski.
\newblock Locally adapted hierarchical basis preconditioning.
\newblock In {\em SIGGRAPH}, 2006.

\bibitem{Vatanen2013}
T.~Vatanen, T.~Raiko, H.~Valpola, and Y.~LeCun.
\newblock Pushing stochastic gradient towards second-order
  methods--backpropagation learning with transformations in nonlinearities.
\newblock In {\em Neural Information Processing}, 2013.

\bibitem{Vedaldi2008}
A.~Vedaldi and B.~Fulkerson.
\newblock {VLFeat}: An open and portable library of computer vision algorithms,
  2008.

\bibitem{Venables1999}
W.~Venables and B.~Ripley.
\newblock Modern applied statistics with s-plus.
\newblock 1999.

\bibitem{Zeiler2014}
M.~D. Zeiler and R.~Fergus.
\newblock Visualizing and understanding convolutional neural networks.
\newblock In {\em ECCV}, 2014.

\end{thebibliography}
}

\newpage

\appendix
\section{Object Detection Baselines}

In this section we introduce our detection method based on the baseline Faster R-CNN \cite{Ren2015} system.
The models are initialized by the ImageNet classification models, and then fine-tuned on the object detection data. We have experimented with ResNet-50/101 at the time of the ILSVRC \& COCO 2015 detection competitions.

Unlike VGG-16 used in \cite{Ren2015}, our ResNet has no hidden fc layers. We adopt the idea of ``Networks on Conv feature maps'' (NoC) \cite{Ren2015a} to address this issue.
We compute the full-image shared conv feature maps using those layers whose strides on the image are no greater than 16 pixels (\ie, conv1, conv2\_ x, conv3\_x, and conv4\_x, totally 91 conv layers in ResNet-101; Table~\ref{tab:arch}). We consider these layers as analogous to the 13 conv layers in VGG-16, and by doing so, both ResNet and VGG-16 have conv feature maps of the same total stride (16 pixels).
These layers are shared by a region proposal network (RPN, generating 300 proposals) \cite{Ren2015} and a Fast R-CNN detection network \cite{Girshick2015}.
RoI pooling \cite{Girshick2015} is performed before conv5\_1. On this RoI-pooled feature, all layers of conv5\_x and up are adopted for each region, playing the roles of VGG-16's fc layers.
The final classification layer is replaced by two sibling layers (classification and box regression \cite{Girshick2015}).

For the usage of BN layers, after pre-training, we compute the BN statistics (means and variances) for each layer on the ImageNet training set. Then the BN layers are fixed during fine-tuning for object detection. As such, the BN layers become linear activations with constant offsets and scales, and BN statistics are not updated by fine-tuning. We fix the BN layers mainly for reducing memory consumption in Faster R-CNN training.

\vspace{.5em}
\noindent\textbf{PASCAL VOC}

Following \cite{Girshick2015,Ren2015}, for the PASCAL VOC 2007 \emph{test} set, we use the 5k \emph{trainval} images in VOC 2007 and 16k \emph{trainval} images in VOC 2012 for training (``07+12''). For the PASCAL VOC 2012 \emph{test} set, we use the 10k \emph{trainval}+\emph{test} images in VOC 2007 and 16k \emph{trainval} images in VOC 2012 for training (``07++12''). The hyper-parameters for training Faster R-CNN are the same as in \cite{Ren2015}.
Table~\ref{tab:detection_voc} shows the results. ResNet-101 improves the mAP by $>$3\% over VGG-16. This gain is solely because of the improved features learned by ResNet.

\vspace{.5em}
\noindent\textbf{MS COCO}

The MS COCO dataset \cite{Lin2014} involves 80 object categories. We evaluate the PASCAL VOC metric (mAP @ IoU = 0.5) and the standard COCO metric (mAP @ IoU = .5:.05:.95). We use the 80k images on the train set for training and the 40k images on the val set for evaluation.
Our detection system for COCO is similar to that for PASCAL VOC.
We train the COCO models with an 8-GPU implementation, and thus the RPN step has a mini-batch size of 8 images (\ie, 1 per GPU) and the Fast R-CNN step has a mini-batch size of 16 images. The RPN step and Fast R-CNN step are both trained for 240k iterations with a learning rate of 0.001 and then for 80k iterations with 0.0001.

Table~\ref{tab:detection_coco} shows the results on the MS COCO validation set. ResNet-101 has a 6\% increase of mAP@[.5, .95] over VGG-16, which is a 28\% relative improvement, solely contributed by the features learned by the better network. Remarkably, the mAP@[.5, .95]'s absolute increase (6.0\%) is nearly as big as mAP@.5's (6.9\%). This suggests that a deeper network can improve both recognition and localization.

\section{Object Detection Improvements}

For completeness, we report the improvements made for the competitions. These improvements are based on deep features and thus should benefit from residual learning.

\renewcommand\arraystretch{1.05}
\setlength{\tabcolsep}{4pt}
\begin{table*}[t]
\begin{center}
\small
\begin{tabular}{l|c|c|c|c}
\hline
training data & \multicolumn{2}{c|}{COCO train} & \multicolumn{2}{c}{COCO trainval} \\
\hline
test data & \multicolumn{2}{c|}{COCO val} & \multicolumn{2}{c}{COCO test-dev}\\
\hline
mAP & ~~~~@.5~~~~ & @[.5, .95] & ~~~~@.5~~~~ & @[.5, .95]\\
\hline
baseline Faster R-CNN (VGG-16)    & 41.5 & 21.2 & \\
baseline Faster R-CNN (ResNet-101)    & 48.4 & 27.2 & \\
~+box refinement &  49.9 & 29.9 & \\
~+context &  51.1 & 30.0 & 53.3 & 32.2 \\
~+multi-scale testing & 53.8 & 32.5 & \textbf{55.7} & \textbf{34.9} \\
\hline
ensemble & & & \textbf{59.0} & \textbf{37.4} \\
\hline
\end{tabular}
\end{center}
\vspace{-.5em}
\caption{Object detection improvements on MS COCO using Faster R-CNN and ResNet-101.}
\vspace{-.5em}
\label{tab:detection_coco_improve}
\end{table*}

\newcolumntype{x}[1]{>{\centering}p{#1pt}}
\newcolumntype{y}{>{\centering}p{16pt}}
\renewcommand{\hl}[1]{\textbf{#1}}
\newcommand{\ct}[1]{\fontsize{6pt}{1pt}\selectfont{#1}}
\renewcommand{\arraystretch}{1.2}
\setlength{\tabcolsep}{1.5pt}
\begin{table*}[t]
\begin{center}
\footnotesize
\vspace{1em}
\resizebox{\linewidth}{!}{
\begin{tabular}{l|x{40}|x{54}|x{20}|yyyyyyyyyyyyyyyyyyyc}
\hline
\ct{system} & net & data & mAP & \ct{areo} & \ct{bike} & \ct{bird} & \ct{boat} & \ct{bottle} & \ct{bus} & \ct{car} & \ct{cat} & \ct{chair} & \ct{cow} & \ct{table} & \ct{dog} & \ct{horse} & \ct{mbike} & \ct{person} & \ct{plant} & \ct{sheep} & \ct{sofa} & \ct{train} & \ct{tv} \\
\hline
\footnotesize baseline & \footnotesize VGG-16 & 07+12 & {73.2} & 76.5 & 79.0 & {70.9} & {65.5} & {52.1} & {83.1} & {84.7} & 86.4 & 52.0 & {81.9} & 65.7 & {84.8} & {84.6} & {77.5} & {76.7} & 38.8 & {73.6} & 73.9 & {83.0} & {72.6}\\
\footnotesize baseline & \footnotesize ResNet-101 & 07+12 & 76.4 & 79.8 & 80.7 & 76.2 & 68.3 & 55.9 & 85.1 & 85.3 & \hl{89.8} & 56.7 & 87.8 & 69.4 & 88.3 & 88.9 & 80.9 & 78.4 & 41.7 & 78.6 & 79.8 & 85.3 & 72.0 \\
\footnotesize baseline+++ & \footnotesize ResNet-101 & COCO+07+12 & \hl{85.6} & \hl{90.0} & \hl{89.6} & \hl{87.8} & \hl{80.8} & \hl{76.1} & \hl{89.9} & \hl{89.9} & {89.6} & \hl{75.5} & \hl{90.0} & \hl{80.7} & \hl{89.6} & \hl{90.3} & \hl{89.1} & \hl{88.7} & \hl{65.4} & \hl{88.1} & \hl{85.6} & \hl{89.0} & \hl{86.8} \\
\hline
\end{tabular}
}
\end{center}
\vspace{-.5em}
\caption{Detection results on the PASCAL VOC 2007 test set. The baseline is the Faster R-CNN system. The system ``baseline+++'' include box refinement, context, and multi-scale testing in Table~\ref{tab:detection_coco_improve}.}
\label{tab:voc07_all}
\begin{center}
\footnotesize
\resizebox{\linewidth}{!}{
\begin{tabular}{l|x{40}|x{54}|x{20}|yyyyyyyyyyyyyyyyyyyc}
\hline
\ct{system} & net & data & mAP & \ct{areo} & \ct{bike} & \ct{bird} & \ct{boat} & \ct{bottle} & \ct{bus} & \ct{car} & \ct{cat} & \ct{chair} & \ct{cow} & \ct{table} & \ct{dog} & \ct{horse} & \ct{mbike} & \ct{person} & \ct{plant} & \ct{sheep} & \ct{sofa} & \ct{train} & \ct{tv} \\
\hline
\footnotesize baseline & \footnotesize VGG-16 & 07++12 & {70.4} & {84.9} & {79.8} & {74.3} & {53.9} & {49.8} & 77.5 & {75.9} & 88.5 & {45.6} & {77.1} & {55.3} & 86.9 & {81.7} & {80.9} & {79.6} & {40.1} & {72.6} & 60.9 & {81.2} & 61.5\\
\footnotesize baseline & \footnotesize ResNet-101 & 07++12 & 73.8 & 86.5 & 81.6 & 77.2 & 58.0 & 51.0 & 78.6 & 76.6 & 93.2 & 48.6 & 80.4 & 59.0 & 92.1 & 85.3 & 84.8 & 80.7 & 48.1 & 77.3 & 66.5 & 84.7 & 65.6 \\
\footnotesize baseline+++ & \footnotesize ResNet-101 & COCO+07++12 & \hl{83.8} & \hl{92.1} & \hl{88.4} & \hl{84.8} & \hl{75.9} & \hl{71.4} & \hl{86.3} & \hl{87.8} & \hl{94.2} & \hl{66.8} & \hl{89.4} & \hl{69.2} & \hl{93.9} & \hl{91.9} & \hl{90.9} & \hl{ 89.6} & \hl{67.9} & \hl{88.2} & \hl{76.8} & \hl{90.3} & \hl{80.0} \\
\hline
\end{tabular}
}
\end{center}
\vspace{-.5em}
\caption{Detection results on the PASCAL VOC 2012 test set (\url{http://host.robots.ox.ac.uk:8080/leaderboard/displaylb.php?challengeid=11&compid=4}). The baseline is the Faster R-CNN system. The system ``baseline+++'' include box refinement, context, and multi-scale testing in Table~\ref{tab:detection_coco_improve}.}
\label{tab:voc12_all}
\end{table*}

\vspace{.5em}
\noindent\textbf{MS COCO}

\noindent\emph{Box refinement.} Our box refinement partially follows the iterative localization in \cite{Gidaris2015}.
In Faster R-CNN, the final output is a regressed box that is different from its proposal box. So for inference, we pool a new feature from the regressed box and obtain a new classification score and a new regressed box.
We combine these 300 new predictions with the original 300 predictions. Non-maximum suppression (NMS) is applied on the union set of predicted boxes using an IoU threshold of 0.3 \cite{Girshick2014}, followed by box voting \cite{Gidaris2015}.
Box refinement improves mAP by about 2 points (Table~\ref{tab:detection_coco_improve}).

\vspace{.5em}
\noindent\emph{Global context.} We combine global context in the Fast R-CNN step. Given the full-image conv feature map, we pool a feature by global Spatial Pyramid Pooling \cite{He2014} (with a ``single-level'' pyramid) which can be implemented as ``RoI'' pooling using the entire image's bounding box as the RoI. This pooled feature is fed into the post-RoI layers to obtain a global context feature. This global feature is concatenated with the original per-region feature, followed by the sibling classification and box regression layers. This new structure is trained end-to-end.
Global context improves mAP@.5 by about 1 point (Table~\ref{tab:detection_coco_improve}).

\vspace{.5em}
\noindent\emph{Multi-scale testing.} In the above, all results are obtained by single-scale training/testing as in \cite{Ren2015}, where the image's shorter side is $s=600$ pixels. Multi-scale training/testing has been developed in \cite{He2014,Girshick2015} by selecting a scale from a feature pyramid, and in \cite{Ren2015a} by using maxout layers. In our current implementation, we have performed multi-scale \emph{testing} following \cite{Ren2015a}; we have not performed multi-scale training because of limited time. In addition, we have performed multi-scale testing only for the Fast R-CNN step (but not yet for the RPN step).
With a trained model, we compute conv feature maps on an image pyramid, where the image's shorter sides are $s\in\{200, 400, 600, 800, 1000\}$. We select two adjacent scales from the pyramid following \cite{Ren2015a}. RoI pooling and subsequent layers are performed on the feature maps of these two scales \cite{Ren2015a}, which are merged by maxout as in \cite{Ren2015a}.
Multi-scale testing improves the mAP by over 2 points (Table~\ref{tab:detection_coco_improve}).

\vspace{.5em}
\noindent\emph{Using validation data.} Next we use the 80k+40k trainval set for training and the 20k test-dev set for evaluation. The test-dev set has no publicly available ground truth and the result is reported by the evaluation server. Under this setting, the results are an mAP@.5 of 55.7\% and an mAP@[.5, .95] of 34.9\% (Table~\ref{tab:detection_coco_improve}). This is our single-model result.

\vspace{.5em}
\noindent\emph{Ensemble.} In Faster R-CNN, the system is designed to learn region proposals and also object classifiers, so an ensemble can be used to boost both tasks. We use an ensemble for proposing regions, and the union set of proposals are processed by an ensemble of per-region classifiers.
Table~\ref{tab:detection_coco_improve} shows our result based on an ensemble of 3 networks. The mAP is 59.0\% and 37.4\% on the test-dev set. \emph{This result won the 1st place in the detection task in COCO 2015.}

\vspace{1em}
\noindent\textbf{PASCAL VOC}

We revisit the PASCAL VOC dataset based on the above model. With the single model on the COCO dataset (55.7\% mAP@.5 in Table~\ref{tab:detection_coco_improve}), we fine-tune this model on the PASCAL VOC sets. The improvements of box refinement, context, and multi-scale testing are also adopted. By doing so we achieve 85.6\% mAP on PASCAL VOC 2007 (Table~\ref{tab:voc07_all}) and 83.8\% on PASCAL VOC 2012 (Table~\ref{tab:voc12_all})\footnote{\fontsize{6.5pt}{1em}\selectfont\url{http://host.robots.ox.ac.uk:8080/anonymous/3OJ4OJ.html}, submitted on 2015-11-26.}. The result on PASCAL VOC 2012 is 10 points higher than the previous state-of-the-art result \cite{Gidaris2015}.

\vspace{1em}
\noindent\textbf{ImageNet Detection}

\renewcommand\arraystretch{1.2}
\setlength{\tabcolsep}{10pt}
\begin{table}[t]
\begin{center}
\small
\begin{tabular}{l|c|c}
\hline
  & val2 & test \\
\hline
GoogLeNet \cite{Szegedy2015} (ILSVRC'14) & - & 43.9 \\
\hline
our single model (ILSVRC'15) & 60.5 & 58.8  \\
our ensemble (ILSVRC'15) & \textbf{63.6} & \textbf{62.1} \\
\hline
\end{tabular}
\end{center}
\vspace{-.5em}
\caption{Our results (mAP, \%) on the ImageNet detection dataset. Our detection system is Faster R-CNN \cite{Ren2015} with the improvements in Table~\ref{tab:detection_coco_improve}, using ResNet-101.
}
\vspace{-.5em}
\label{tab:imagenet_det}
\end{table}

The ImageNet Detection (DET) task involves 200 object categories. The accuracy is evaluated by mAP@.5.
Our object detection algorithm for ImageNet DET is the same as that for MS COCO in Table~\ref{tab:detection_coco_improve}. The networks are pre-trained on the 1000-class ImageNet classification set, and are fine-tuned on the DET data. We split the validation set into two parts (val1/val2) following \cite{Girshick2014}. We fine-tune the detection models using the DET training set and the val1 set. The val2 set is used for validation. We do not use other ILSVRC 2015 data. Our single model with ResNet-101 has 58.8\% mAP and our ensemble of 3 models has 62.1\% mAP on the DET test set (Table~\ref{tab:imagenet_det}). \emph{This result won the 1st place in the ImageNet detection task in ILSVRC 2015}, surpassing the second place by \textbf{8.5 points} (absolute).


\section{ImageNet Localization}
\label{sec:appendix_localization}

\renewcommand\arraystretch{1.05}
\setlength{\tabcolsep}{2pt}
\begin{table}[t]
\begin{center}
\small
\resizebox{1.0\linewidth}{!}{
\begin{tabular}{c|c|c|c|c|c}
\hline
\tabincell{c}{LOC \\ method} & \tabincell{c}{LOC \\ network} & testing & \tabincell{c}{LOC error \\on GT CLS} & \tabincell{c}{classification\\ network} & \tabincell{c}{top-5 LOC error \\ on predicted CLS} \\
\hline
VGG's \cite{Simonyan2015} & VGG-16 & 1-crop & 33.1 \cite{Simonyan2015} & & \\
RPN & ResNet-101 & 1-crop & 13.3 & & \\
RPN & ResNet-101 & dense & 11.7 & & \\
\hline
RPN & ResNet-101 & dense & & ResNet-101 & 14.4 \\
RPN+RCNN & ResNet-101 & dense & & ResNet-101 & \textbf{10.6} \\
RPN+RCNN & ensemble & dense & & ensemble & \textbf{8.9} \\
\hline
\end{tabular}
}
\end{center}
\vspace{-.5em}
\caption{Localization error (\%) on the ImageNet validation. In the column of ``LOC error on GT class'' (\cite{Simonyan2015}), the ground truth class is used.
In the ``testing'' column, ``1-crop'' denotes testing on a center crop of 224$\times$224 pixels, ``dense'' denotes dense (fully convolutional) and multi-scale testing.
}
\vspace{-.5em}
\label{tab:localization}
\end{table}

The ImageNet Localization (LOC) task \cite{Russakovsky2014} requires to classify and localize the objects.
Following \cite{Sermanet2014,Simonyan2015}, we assume that the image-level classifiers are first adopted for predicting the class labels of an image, and the localization algorithm only accounts for predicting bounding boxes based on the predicted classes. We adopt the ``per-class regression'' (PCR) strategy \cite{Sermanet2014,Simonyan2015}, learning a bounding box regressor for each class. We pre-train the networks for ImageNet classification and then fine-tune them for localization.
We train networks on the provided 1000-class ImageNet training set.

Our localization algorithm is based on the RPN framework of \cite{Ren2015} with a few modifications.
Unlike the way in \cite{Ren2015} that is category-agnostic, our RPN for localization is designed in a \emph{per-class} form. This RPN ends with two sibling 1$\times$1 convolutional layers for binary classification (\emph{cls}) and box regression (\emph{reg}), as in \cite{Ren2015}. The \emph{cls} and \emph{reg} layers are both in a \emph{per-class} from, in contrast to \cite{Ren2015}. Specifically, the \emph{cls} layer has a 1000-d output, and each dimension is \emph{binary logistic regression} for predicting being or not being an object class; the \emph{reg} layer has a 1000$\times$4-d output consisting of box regressors for 1000 classes.
As in \cite{Ren2015}, our bounding box regression is with reference to multiple translation-invariant ``anchor'' boxes at each position.

As in our ImageNet classification training (Sec.~\ref{sec:impl}), we randomly sample 224$\times$224 crops for data augmentation. We use a mini-batch size of 256 images for fine-tuning.
To avoid negative samples being dominate, 8 anchors are randomly sampled for each image, where the sampled positive and negative anchors have a ratio of 1:1 \cite{Ren2015}. For testing, the network is applied on the image fully-convolutionally.

\renewcommand\arraystretch{1.05}
\setlength{\tabcolsep}{10pt}
\begin{table}[t]
\begin{center}
\small
\begin{tabular}{l|c|c}
\hline
  \multicolumn{1}{c|}{\multirow{2}{*}{method}}  & \multicolumn{2}{c}{top-5 localization err} \\\cline{2-3}
  & val & test \\
\hline
OverFeat \cite{Sermanet2014} (ILSVRC'13) & 30.0 & 29.9 \\
GoogLeNet \cite{Szegedy2015} (ILSVRC'14) & - & 26.7 \\
VGG \cite{Simonyan2015} (ILSVRC'14) & 26.9 & 25.3 \\
\hline
ours (ILSVRC'15) & \textbf{8.9} & \textbf{9.0} \\
\hline
\end{tabular}
\end{center}
\vspace{-.5em}
\caption{Comparisons of localization error (\%) on the ImageNet dataset with state-of-the-art methods.
}
\vspace{-.5em}
\label{tab:localization_all}
\end{table}

Table~\ref{tab:localization} compares the localization results. Following \cite{Simonyan2015}, we first perform ``oracle'' testing using the ground truth class as the classification prediction. VGG's paper \cite{Simonyan2015} reports a center-crop error of 33.1\% (Table~\ref{tab:localization}) using ground truth classes. Under the same setting, our RPN method using ResNet-101 net significantly reduces the center-crop error to 13.3\%. This comparison demonstrates the excellent performance of our framework.
With dense (fully convolutional) and multi-scale testing, our ResNet-101 has an error of 11.7\% using ground truth classes. Using ResNet-101 for predicting classes (4.6\% top-5 classification error, Table~\ref{tab:single}), the top-5 localization error is 14.4\%.

The above results are only based on the \emph{proposal network} (RPN) in Faster R-CNN \cite{Ren2015}. One may use the \emph{detection network} (Fast R-CNN \cite{Girshick2015}) in Faster R-CNN to improve the results. But we notice that on this dataset, one image usually contains a single dominate object, and the proposal regions highly overlap with each other and thus have very similar RoI-pooled features. As a result, the image-centric training of Fast R-CNN \cite{Girshick2015} generates samples of small variations, which may not be desired for stochastic training. Motivated by this, in our current experiment we use the original R-CNN \cite{Girshick2014} that is RoI-centric, in place of Fast R-CNN.

Our R-CNN implementation is as follows. We apply the per-class RPN trained as above on the training images to predict bounding boxes for the ground truth class. These predicted boxes play a role of class-dependent proposals.
For each training image, the highest scored 200 proposals are extracted as training samples to train an R-CNN classifier. The image region is cropped from a proposal, warped to 224$\times$224 pixels, and fed into the classification network as in R-CNN \cite{Girshick2014}. The outputs of this network consist of two sibling fc layers for \emph{cls} and \emph{reg}, also in a per-class form.
This R-CNN network is fine-tuned on the training set using a mini-batch size of 256 in the RoI-centric fashion. For testing, the RPN generates the highest scored 200 proposals for each predicted class, and the R-CNN network is used to update these proposals' scores and box positions.

This method reduces the top-5 localization error to 10.6\% (Table~\ref{tab:localization}). This is our single-model result on the validation set. Using an ensemble of networks for both classification and localization, we achieve a top-5 localization error of 9.0\% on the test set. This number significantly outperforms the ILSVRC 14 results (Table~\ref{tab:localization_all}), showing a 64\% relative reduction of error. \emph{This result won the 1st place in the ImageNet localization task in ILSVRC 2015.}

\end{document}